\documentclass[12pt]{iopart}
\usepackage{iopams}

\usepackage{bm}
\usepackage{amsthm,amsfonts,epsfig,setspace}
\usepackage{colortbl,color}
\usepackage[margin=1in]{geometry}
\usepackage{epstopdf}
\usepackage{amssymb}
\usepackage{mathrsfs}
\usepackage{latexsym,bm}
\usepackage{amsthm}
\usepackage{float}
\usepackage{graphicx}
\usepackage{CJK}
\usepackage{graphics}
\usepackage{color}
\usepackage{subfigure}
\usepackage[english]{babel}

\usepackage{algorithm}
\usepackage{algpseudocode}

\usepackage{booktabs}
\usepackage{diagbox,multirow,arydshln}
\usepackage{array}
\usepackage{hhline}
\usepackage{makecell}
\usepackage{float}
\usepackage{framed}
\usepackage{varwidth}
\usepackage[export]{adjustbox}
\usepackage{colortbl}
\usepackage{sidecap}
\usepackage[section]{placeins}
\usepackage{framed}
\usepackage{caption}
\usepackage{placeins}
\usepackage{needspace}

\graphicspath{ {figures/} }

\begin{document}

\title[ResDynUNet++ for dual-spectral CT]{ResDynUNet++: A nested U-Net with residual dynamic convolution blocks for dual-spectral CT}

\author{Ze Yuan$^{1}$, Wenbin Li$^{1,\ast}$, Shusen Zhao$^{2,3}$}

\address{$^{1}$ School of Science, Harbin Institute of Technology, Shenzhen, Shenzhen, 518055, China}
\address{$^{2}$ National Center for Applied Mathematics Shenzhen, Southern University of Science and Technology, Shenzhen 518055, China}
\address{$^{3}$ Detection Institute for Advanced Technology Longhua-Shenzhen, Shenzhen 518000, China}
\address{$^{\ast}$ Corresponding author}
\ead{25B958001@stu.hit.edu.cn (Ze Yuan), liwenbin@hit.edu.cn (Wenbin Li), zhaoss@sustech.edu.cn (Shusen Zhao)}
	
	\begin{abstract}
	We propose a hybrid reconstruction framework for dual-spectral CT (DSCT) that integrates iterative methods with deep learning models. The reconstruction process consists of two complementary components: a knowledge-driven module and a data-driven module. In the knowledge-driven phase, we employ the oblique projection modification technique (OPMT) to reconstruct an intermediate solution of the basis material images from the projection data. We select OPMT for this role because of its fast convergence, which allows it to rapidly generate an intermediate solution that successfully achieves basis material decomposition. Subsequently, in the data-driven phase, we introduce a novel neural network, ResDynUNet++, to refine this intermediate solution. The ResDynUNet++ is built upon a UNet++ backbone by replacing standard convolutions with residual dynamic convolution blocks, which combine the adaptive, input-specific feature extraction of dynamic convolution with the stable training of residual connections. This architecture is designed to address challenges like channel imbalance and near-interface large artifacts in DSCT, producing clean and accurate final solutions. Extensive experiments on both synthetic phantoms and real clinical datasets validate the efficacy and superior performance of the proposed method.
	\end{abstract}

	\maketitle
	
	
	\section{Introduction}
	
	Computerized tomography (CT) is an indispensable tool in modern medicine, providing detailed cross-sectional images crucial for diagnosis and treatment planning. However, conventional single-energy CT suffers from intrinsic limitations. As the X-ray attenuation coefficient depends on both the atomic number and photon energy, tissues with distinct compositions may exhibit identical attenuation values, hindering differentiation \cite{nat01}. Additionally, the polychromatic nature of the X-ray beam frequently leads to artifacts, with beam hardening being a typical example \cite{brodi76,vanetc11}.	
	
	Dual-spectral CT (DSCT), frequently referred to as dual-energy CT (DECT) \cite{joh12}, represents a paradigm shift in tomographic imaging. Unlike conventional single-energy systems, DSCT exploits the energy dependence of the linear attenuation coefficient by acquiring projection data at two distinct X-ray spectra. This spectral separation enables the decomposition of the scanned object into two constituent basis materials, such as bone and soft tissue, fundamentally characterizing the contributions of the photoelectric effect and Compton scattering \cite{alvmac76}. The resulting material density images yield significant clinical advantages, including precise material differentiation, the synthesis of virtual monochromatic images \cite{yuchretc11} to optimize contrast-to-noise ratios, and the substantial mitigation of beam-hardening artifacts \cite{colsin85}, ultimately providing superior quantitative diagnostic information.
	
	Reconstructing images from DSCT data is a complex inverse problem. Conventional methods largely rely on filtered back-projection (FBP) algorithm to transform the projection data back to a reconstruction in the spatial domain \cite{kat04,pansidvan09}. For example, in image-domain decomposition methods \cite{wuetc20}, FBP is used to reconstruct two independent images from the high- and low-energy projection data, and the basis material images are then decomposed from these two recovered images. In projection-domain decomposition methods \cite{chuhua88}, the projection data are first decomposed into equivalent basis material projections, and FBP is then employed to reconstruct basis material images from the decomposed data.
The direct FBP-based methods are valuable for their simplicity and speed in clinical practice, but FBP is sensitive to noise so that the inversion results rely heavily on the completeness and high quality of the measurement data \cite{aoliqia21}.

	With the rapid development of computing hardware and reconstruction algorithms, iterative methods for DSCT have gained popularity \cite{sukcli00,elbfes02,zhazhazha14,zhapanzhaxiazha21,gaopanche22}. These methods formulate reconstruction as a unified optimization problem, seeking basis material images from projection data by solving a large system of model equations. For example, the extended algebraic reconstruction technique (E-ART) \cite{zhazhazha14}, a prominent iterative algorithm for DSCT, extends the classic ART method \cite{gorbenher70} to address the nonlinear system modeling DSCT reconstruction. E-ART can produce high-quality basis material images, especially from sparse-angle data, but it is computationally demanding and often suffers from slow convergence. As a more recent and efficient alternative, the oblique projection modification technique (OPMT) is introduced \cite{zhapanzhaxiazha21}. OPMT calculates an oblique projection path to model and compensate for physical shifts between sequential high- and low-energy scans, thereby effectively reducing decomposition artifacts and yielding much faster convergence.
	
	In parallel to these developments, the rise of deep learning, particularly convolutional neural networks (CNNs), has achieved significant success in medical image reconstruction. By learning intricate patterns from large datasets, these models have shown an extraordinary ability to suppress noise, eliminate artifacts, and recover fine structural details. The U-Net \cite{ronneberger2015u,jiaetc25}, with its elegant symmetric encoder-decoder design and skip connections, quickly became a foundational architecture. Its strength lies in preserving multi-scale features, which is essential for accurate image restoration. UNet++ further refined this concept by introducing nested and dense skip connections \cite{zhou2018unet++}. This design shortens the pathway for information to flow between the encoder and decoder, enabling more effective feature fusion at different scales and leading to superior performance on challenging imaging tasks.
	
	In this work, we propose a hybrid reconstruction framework for DSCT that integrates iterative methods with deep learning models. The reconstruction process consists of a knowledge-driven part and a data-driven part \cite{aoliqia21,adlokt17,adlokt18,jinmccfrouns17}. In the knowledge-driven part, an iterative algorithm is employed to reconstruct an intermediate solution of the basis material images from the projection data. In the data-driven part, a deep neural network is developed to refine the intermediate solution, removing artifacts due to data noise and the intrinsic limitations of the iterative algorithm. We select OPMT algorithm as the model-driven part of the reconstruction framework. Because of its fast convergence, the OPMT can rapidly generate an intermediate solution that successfully achieves basis material decomposition, a task that remains challenging for purely data-driven approaches. Then we propose a neural network, named ResDynUNet++, as the data-driven part of the reconstruction framework. The architecture of ResDynUNet++ is designed to address challenges like channel imbalance and near-interface large artifacts in DSCT, producing clean and more accurate final solutions.
	The hybrid framework  preserves the mathematical formulation of the DSCT model, and is able to capture the latent features of the projection data, yielding a data and knowledge driven reconstruction.
		
	The remainder of this paper is structured as follows. Section 2 formulates the inverse problem of DSCT reconstruction. Section 3 details the proposed methodology, presenting the data and knowledge driven hybrid reconstruction framework, describing the OPMT algorithm, and introducing the ResDynUNet++ network architecture. Section 4 presents and analyzes the experimental results from both synthetic and clinical datasets. Finally, Section 5 draws the conclusion.

	\section{Inverse problem of dual-spectral CT (DSCT)}
	
\begin{figure}[htbp!]
		\centering
		\includegraphics[scale=0.32,angle=0]{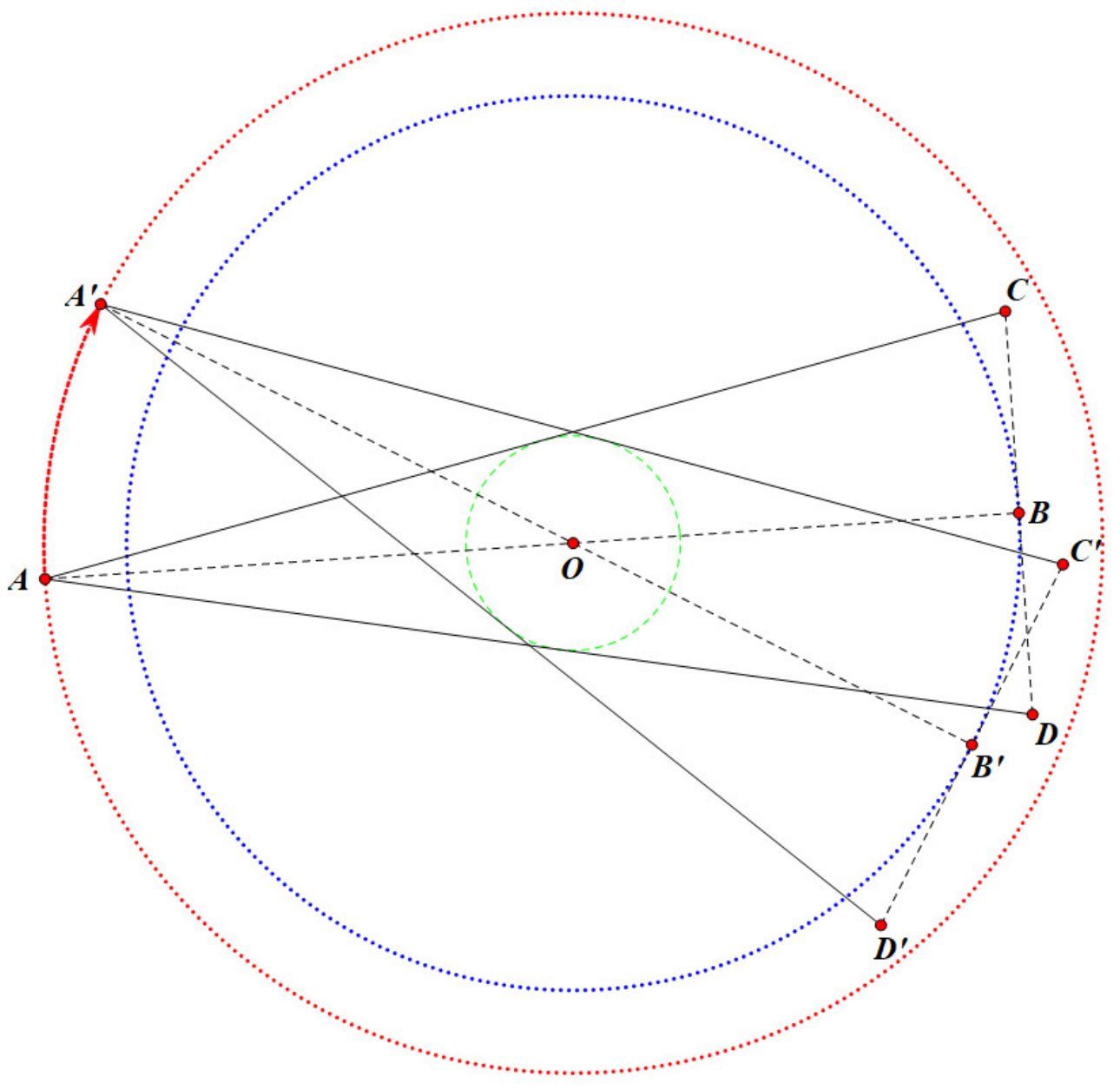}
		\caption{Geometry of a fan-beam CT system.}
		\label{fig_fanbeam}
	\end{figure}

	\Fref{fig_fanbeam} shows the geometry of a fan-beam CT system, where $A$ and $A'$ illustrate the rotated X-ray source, $CD$ and $C'D'$ illustrate the rotated line of detectors, and $O$ is the center of rotation. The distance from the source to the center of rotation is the Source-to-Object Distance (SOD), denoted as $D_1$. The distance from the center of rotation to the detector array is the Object-to-Detector Distance (ODD), denoted as $D_2$. The circular region consistently irradiated by the fan beam from all projection angles defines the Field of View (FOV), as shown by the green circle in \Fref{fig_fanbeam}. Denoting $L_H$ as the half-length of the detector array, the FOV radius is given by $r = \frac{D_1 \cdot L_H}{\sqrt{L_H^2 + (D_1 + D_2)^2}}$.
	
	For a DSCT system with two distinct X-ray spectra, $S_1(E)$ and $S_2(E)$, the projection data $p_k(L)$ for a given X-ray path $L$ is modeled as
\begin{equation}
p_k(L) = -\ln \left( \int S_k(E) e^{-\int_L \mu(x,E) \rmd l} \rmd E \right),\quad k=1,2\,, \label{eq:projection}
\end{equation}
where $\mu(x,E)$ represents the linear attenuation coefficient at position $x$ and energy $E$. Given $S_k(E)\ (k=1,2)$, the inverse problem of DSCT aims to solve $\mu(x,E)$ from the projection data $p_k(L)$, $\forall$ $k=1,2$ and $L\in \Pi_L$; $\Pi_L$ denotes the index set of the X-ray paths. 
In this work, we consider the following decomposition for $\mu(x,E)$ \cite{zhazhazha14,lizhazha18},
\begin{equation}
 \mu(x,E) = \phi(E)f(x) + \theta(E)g(x)\,, \label{eq:mu}
\end{equation}
where $f(x)$ and $g(x)$ represent the mass densities of two selected basis materials, e.g., bone and water, and $\phi(E)$ and $\theta(E)$ are their respective mass attenuation coefficients. With the pre-defined coefficients $\phi(E)$ and $\theta(E)$, the inverse problem reduces to solving the density functions $f(x)$ and $g(x)$.
	
	Consider the discrete form of equation (\ref{eq:projection}). Given the number of projection angles $n_S$ and the number of detector elements $n_D$, the total number of X-ray paths is $n_S n_D$. Let $\mathbf{f}$ and $\mathbf{g}$ denote the flattened vectors of the discretized density functions $f(x)$ and $g(x)$, respectively,
	\begin{equation}
		\mathbf{f} = ( f_1 , f_2, \cdots , f_{N_R} ),\quad  \mathbf{g} = ( g_1 , g_2, \cdots , g_{N_R} )\,,
	\end{equation}
where $N_R = n_{R} \times n_{R}$ is the number of pixels in each discretized density image.  Then we introduce a projection matrix $R\in\mathbb{R}^{n_S n_D \times N_R}$ that maps the density image to the projection domain: $R=(r_{ij})_{n_S n_D \times N_R}$, where $r_{ij}$ represents the contribution of the $j$-th pixel of $\mathbf{f}$ or $\mathbf{g}$ to the projection along the $i$-th X-ray path. 
Let $R_l$ denote the $l$-th row of the projection matrix $R$. Divide the valid energy range of the $k$-th X-ray spectrum	into $M_k$ parts with subinterval length $\delta_{E}$, and denote $S_{k,m}$, $\phi_m$ and $\theta_m$ as the sampling values of $S_k(E)$, $\phi(E)$ and $\theta(E)$ in the $m$-th subinterval. The discrete form of equation (\ref{eq:projection}) reads as follows,
	\begin{equation}
		p_{k,l} = -\ln \left( \sum_{m=1}^{M_k} S_{k,m} \delta_{E} e^{-\phi_m R_l \mathbf{f} - \theta_m R_l \mathbf{g}} \right),\quad k=1,2,\quad l=1,2,\cdots, n_S n_D.
		\label{eq:d_projection}
	\end{equation}
The discrete inverse problem is to solve the density image vectors $\mathbf{f}$ and $\mathbf{g}$ from the projection data $\mathbf{p}_k: = (p_{k,l})_{1\le l\le n_S n_D}$, $\forall\,k=1,2$.

	\section{Methodology}
	
	\subsection{A data and knowledge driven reconstruction framework}

	We propose a hybrid framework that is both knowledge-driven and data-driven \cite{aoliqia21, adlokt17} for the inverse problem of DSCT. This approach combines a classical iterative algorithm, which incorporates the physical model knowledge, with a deep learning network that learns from data to refine the solution.
	
	Let $\mathcal{F}$ represent the operator for a single iteration of the selected iterative algorithm; in this work, we will consider the oblique projection modification technique (OPMT) \cite{zhapanzhaxiazha21}. After $n$ iterations, the intermediate solution $(\mathbf{\tilde{f}}, \mathbf{\tilde{g}})$ is obtained from the projection data $\mathbf{p}$:
	\begin{equation}  \label{eqn5}
		(\mathbf{\tilde{f}}, \mathbf{\tilde{g}}) = \mathcal{F}^n(\mathbf{p}),
	\end{equation}
	where $\mathcal{F}^n = \mathcal{F} \circ \dots \circ \mathcal{F}$ denotes applying the operator $n$ times. This first stage is the knowledge-driven component, as it directly utilizes the mathematical model of the DSCT forward projection to produce a physically plausible solution. This intermediate solution is typically suboptimal, primarily due to data noise and the intrinsic limitations of the iterative algorithm.
	
	Next, let $\Lambda_{\Theta}$ denote the operator of our proposed deep neural network, ResDynUNet++, with $\Theta$ being the set of trainable network parameters. This network takes the intermediate solution as input and produces the final, refined image. The complete reconstruction operator, $\mathcal{A}_{\Theta}^{\dagger}$, can thus be expressed as the composition of these two stages:
	\begin{equation} \label{eq:Full_progress}
		(\mathbf{f}, \mathbf{g}) = \mathcal{A}_{\Theta}^{\dagger}(\mathbf{p}) := \Lambda_{\Theta} \circ \mathcal{F}^n (\mathbf{p}) = \Lambda_{\Theta}(\mathbf{\tilde{f}}, \mathbf{\tilde{g}}).
	\end{equation}
	The second stage constitutes the data-driven component. The network $\Lambda_{\Theta}$ is not explicitly programmed with the physics of the system. Instead, it learns a complex mapping from noisy inputs to clean outputs through supervised training on a large dataset. This enables the correction of artifacts and noise patterns that are difficult to model analytically.
	
	The hybrid reconstruction framework allows the OPMT algorithm to efficiently handle the core physics-based inversion, while the deep network focuses on the sophisticated task of image quality enhancement by leveraging features learned from data.

	\subsection{OPMT algorithm}
	We consider the oblique projection modification technique (OPMT) \cite{zhapanzhaxiazha21} for constructing the iteration operator $\mathcal{F}$ in formulas (\ref{eqn5}) and (\ref{eq:Full_progress}). Comparing to typical approaches like E-ART \cite{zhazhazha14}, which often takes hundreds of iterations to separate the basis material images $\mathbf{f}$ and $\mathbf{g}$, the OPMT algorithm accelerates the convergence speed to efficiently achieve the intermediate solution $(\mathbf{\tilde{f}}, \mathbf{\tilde{g}})$. The trade-off is that the OPMT algorithm is often more sensitive to data noise, yielding an imperfect solution corrupted by artifacts and noise patterns, which can be refined by the subsequent deep learning network. The OPMT algorithm is well-suited to the hybrid reconstruction framework for DSCT because the decomposition of basis materials $\mathbf{f}$ and $\mathbf{g}$ relies on model knowledge, which OPMT can efficiently and rapidly accomplish. Conversely, mitigating noise contamination and enhancing image quality are tasks ideally handled by the deep neural network. In the following, we provide a brief description of the OPMT algorithm for dual-spectral CT. 
	
	Performing a first-order Taylor expansion of equation \eref{eq:d_projection} around the current iterative state $(\mathbf{f}^{(n)}, \mathbf{g}^{(n)})$ yields the linearized system: 
\begin{equation} \label{eqn7}
\frac{\Phi_{k,l}^{(n)}}{q_{k,l}^{(n)}}R_l(\mathbf{f} - \mathbf{f}^{(n)}) + \frac{\Theta_{k,l}^{(n)}}{q_{k,l}^{(n)}} R_l(\mathbf{g} - \mathbf{g}^{(n)}) = p_{k,l} - p_{k,l}^{(n)}, \quad k=1,2,\ \  l=1,\cdots, n_S n_D.
\end{equation}
	In equation \eref{eqn7}, $p_{k,l}$ denotes measured projection data, and
	\begin{eqnarray}
		p_{k,l}^{(n)} = -\ln \sum_{m=1}^{M_k} S_{k,m} \delta_{E} e^{-\phi_{m} R_l \mathbf{f}^{(n)} - \theta_{m} R_l \mathbf{g}^{(n)}}\,, \\
		q_{k,l}^{(n)} = \sum_{m=1}^{M_k} S_{k,m} \delta_{E} e^{-\phi_{m} R_l \mathbf{f}^{(n)} - \theta_{m} R_l \mathbf{g}^{(n)}}\,, \\
		\Phi_{k,l}^{(n)} = \sum_{m=1}^{M_k} S_{k,m} \delta_{E} \phi_{m} e^{-\phi_{m} R_l \mathbf{f}^{(n)} - \theta_{m} R_l \mathbf{g}^{(n)}}\,, \\
		\Theta_{k,l}^{(n)} = \sum_{m=1}^{M_k} S_{k,m} \delta_{E} \theta_{m} e^{-\phi_{m} R_l \mathbf{f}^{(n)} - \theta_{m} R_l \mathbf{g}^{(n)}}\,.
	\end{eqnarray}
For every $l$, equation \eref{eqn7} is a system of linear equations representing two hyperplanes, $H_1$ and $H_2$,
\begin{equation} \label{eq:H1}
\left\{
	\begin{array}{cc}
		H_1:&  a_{11}x_1 + a_{12}x_2 = b_1 \\
		H_2:&  a_{21}x_1 + a_{22}x_2 = b_2
	\end{array}
\right.\,,
\end{equation}
where
	\begin{eqnarray}
		a_{k1} = \frac{\Phi_{k,l}^{(n)}}{q_{k,l}^{(n)}}, \quad a_{k2} = \frac{\Theta_{k,l}^{(n)}}{q_{k,l}^{(n)}},\quad k=1,2 \,, \\
		x_1 = R_l \mathbf{f}, \quad x_2 = R_l \mathbf{g} \,, \\
		b_k = p_{k,l} - p_{k,l}^{(n)} + \frac{\Phi_{k,l}^{(n)}}{q_{k,l}^{(n)}} R_l \mathbf{f}^{(n)} + \frac{\Theta_{k,l}^{(n)}}{q_{k,l}^{(n)}} R_l \mathbf{g}^{(n)}, \quad k=1,2 \,.
	\end{eqnarray}
	To derive $(\mathbf{f}^{(n+1)}, \mathbf{g}^{(n+1)})$ from the current iterative state $(\mathbf{f}^{(n)}, \mathbf{g}^{(n)})$, a natural approach is to first project $(\mathbf{f}^{(n)}, \mathbf{g}^{(n)})$ onto the hyperplane $H_1$ defined in (\ref{eq:H1}), and then project the resulting point onto $H_2$, which is the method of E-ART \cite{zhazhazha14}. However, using data from only one spectrum per projection limits algorithmic efficiency. Convergence can be accelerated by redesigning the projection direction to incorporate data from both spectra simultaneously.
	
	Let $\mathrm{dir}_1$ be the unit normal direction of the hyperplane $H_1$:
	\begin{equation}
		\mathrm{dir}_1 = \frac{(a_{11}, a_{12})}{\sqrt{a_{11}^2 + a_{12}^2}}\,.
	\end{equation}
Considering the direction orthogonal to the normal vector of $H_2$, which can be $(a_{22}, -a_{21})$ or $(-a_{22}, a_{21})$, we choose the one that forms an acute angle with $\mathrm{dir}_1$ and normalize it to define $\mathrm{dir}_2$:
	\begin{equation}
		\mathrm{dir}_2 = \cases{ \frac{(a_{22}, -a_{21})}{\sqrt{a_{21}^2+a_{22}^2}} & if $a_{11}a_{22} > a_{12}a_{21}$ \\ \frac{(-a_{22}, a_{21})}{\sqrt{a_{21}^2+a_{22}^2}} & if $a_{11}a_{22} < a_{12}a_{21}$ }
		\,.
	\end{equation}
The modified projection direction to $H_1$ is then designed as a linear combination of $\mathrm{dir}_1$ and $\mathrm{dir}_2$,
	\begin{equation}
		\mathrm{dir} = \lambda_1 \mathrm{dir}_1 + \lambda_2 \mathrm{dir}_2 \,,
	\end{equation}
where $\lambda_1=\lambda_2=1$ is selected following \cite{zhapanzhaxiazha21}. The resulting iterative formula is
	\begin{equation}
		\pmatrix{ R_l \mathbf{f}^{(n+1)} \cr R_l \mathbf{g}^{(n+1)} } = \pmatrix{ R_l \mathbf{f}^{(n)} \cr R_l \mathbf{g}^{(n)} } + \frac{p_{1,l} - p_{1,l}^{(n)}}{\langle(a_{11}, a_{12}), \mathrm{dir}\rangle} \mathrm{dir}^T\,,
	\end{equation}
which implies that
	\begin{equation}
		\pmatrix{ \mathbf{f}^{(n+1)} \cr \mathbf{g}^{(n+1)} } = \pmatrix{ \mathbf{f}^{(n)} \cr \mathbf{g}^{(n)} } + R_l^{-1} \frac{p_{1,l} - p_{1,l}^{(n)}}{\langle(a_{11}, a_{12}), \mathrm{dir}\rangle} \mathrm{dir}^T\,.
	\end{equation}
Note that when $\lambda_1=1$ and $\lambda_2=0$, the iterative formula is identical to E-ART. The overall process completes by subsequently projecting the result onto $H_2$ using an obliquely selected direction in an analogous manner.
	
	\subsection{Proposed network: ResDynUNet++}
	In this part, we explain the development of our deep neural network, named ResDynUNet++, for the data-driven part $\Lambda_{\Theta}$ in the hybrid reconstruction framework. The proposed network aims to refine the intermediate solution, $(\mathbf{\tilde{f}}, \mathbf{\tilde{g}})$, obtained by the OPMT algorithm. Its main task is to remove artifacts and noise patterns from the intermediate solution of the two basis-material densities, while preserving their physical features in the reconstruction. 

\subsubsection{Challenge 1:  Channel imbalance and overfitting.}
In dual-spectral CT, the network $\Lambda_{\Theta}:\,(\mathbf{\tilde{f}}, \mathbf{\tilde{g}})\to (\mathbf{f}, \mathbf{g})$ requires two input channels and two output channels, each for one of the basis materials.
The disparate nature of the two channels can lead to unbalanced convergence and severe overfitting of the network.
The overfitting occurs when the model learns statistical noise specific to the training set instead of the general underlying features, resulting in a training error that is deceptively low compared to its high generalization error on new data. When using a validation set, it is indicated by a continued decrease in training loss while the validation loss begins to rise, resulting in a persistent expansion of the generalization gap. For dual-spectral CT, we observe that the data-driven network $\Lambda_{\Theta}$ can exhibit severe overfitting in one channel (e.g., for $\mathbf{g}$), while the other channel remains under-converged.
This is typically attributed to channel imbalance, where disparities in properties like noise intensity and pixel-value range (e.g., the maximum pixel value in one channel being much larger than in the other) cause the learning process to be dominated by the channel with higher-magnitude signals. This imbalance prevents the network from learning coherently from both inputs. 
To address it, we initially explored several conventional techniques: adding L1 or L2 regularization terms to the loss function, applying gradient clipping, weighting the loss components for each basis material \cite{chen2018gradnorm}, and even bifurcating the decoder into two parallel paths. However, none of these modifications produced a significant improvement.

As conventional approaches like regularization and loss weighting were insufficient, we realized that the underlying issue lay in the network architecture rather than parameter tuning. To mitigate the problem of channel imbalance, we adopt an architecture from UNet++, whose nested skip pathways effectively bridge the semantic gap between encoder and decoder, thereby promoting more balanced and harmonized feature learning from the heterogeneous input channels. 
At the same time, to combat overfitting, we leverage the deep supervision mechanism intrinsic to UNet++, which enforces feature learning at multiple semantic levels and introduces a built-in form of regularization.

\subsubsection{Challenge 2: Artifacts at interfaces.}
Another challenge in the development of $\Lambda_{\Theta}$ is that the neural network consistently produces large artifacts near interface regions. It tends to blur interface structures and amplify noise artifacts near the interfaces. To address this, we investigated a wide range of potential solutions. 
An initial attempt to augment the loss function with an edge-detection term (e.g., a Sobel operator) was unsuccessful, likely because such operators lacked sufficient contextual awareness. We then explored attention mechanisms, but integrating simple attention blocks like CBAM \cite{woo2018cbam} proved ineffective. Pivoting to models inspired by Vision Transformers (ViTs) \cite{dosovitskiy2020image, lee2021vitgan} to leverage their self-attention mechanism introduced conspicuous grid-like artifacts. Subsequently, we reframed the problem as an image generation task, employing a Wasserstein GAN \cite{arjovsky2017wasserstein} and experimenting with various critic architectures, from standard CNNs to ViTs. While a ViT-based critic showed a marginal advantage, the results still fell short of our requirements. These investigations highlighted the need for a more nuanced mechanism to handle feature extraction, particularly at interface regions.

In this paper, we integrate Dynamic Convolution into our architecture to enhance feature adaptivity. This technique employs a specialized attention mechanism to generate sample-specific convolution kernels, effectively tailoring feature extraction to the unique characteristics of each input. Crucially, the mechanism is spatially aware, allowing the model to apply selective focus to different regions within a single sample.

\subsubsection{ResDynUNet++ architecture and training.}

\paragraph{Backbone: UNet++.}
The U-Net architecture, with its iconic encoder-decoder structure and skip connections, is a cornerstone of medical image processing. We select UNet++, an advanced variant proposed by Zhou \etal \cite{zhou2018unet++}, as our network backbone. Its nested and dense skip pathways are designed to bridge the semantic gap between the encoder and decoder, enabling more effective fusion of features from different semantic levels and improving performance on complex image-to-image tasks. Therefore, UNet++ serves as a concise and effective backbone for addressing the problems mentioned in Challenge 1.

\paragraph{Dynamic convolution.}
The concept of making convolutional kernels input-dependent, rather than using static filters, has evolved through several key works. An early approach is the Dynamic Filter Network, where filters are not learned directly but are generated dynamically by an auxiliary network conditioned on the input \cite{jia2016dynamic}. This idea is then refined for greater efficiency and model capacity with Conditionally Parameterized Convolutions (CondConv) \cite{yang2019condconv}. CondConv learns a set of specialized `expert' kernels and computes sample-specific weights to linearly combine them, thereby improving performance without a commensurate increase in inference cost. Building upon this, \cite{chen2020dynamic} formalizes the kernel aggregation process through an attention mechanism, proposing Dynamic Convolution. This method employs a lightweight attention module to determine the optimal weights for combining multiple parallel kernels into a single, input-specific dynamic kernel for feature extraction.

\begin{figure}[htbp!]
	\centering
	(a){{\includegraphics[scale=0.5,angle=0]{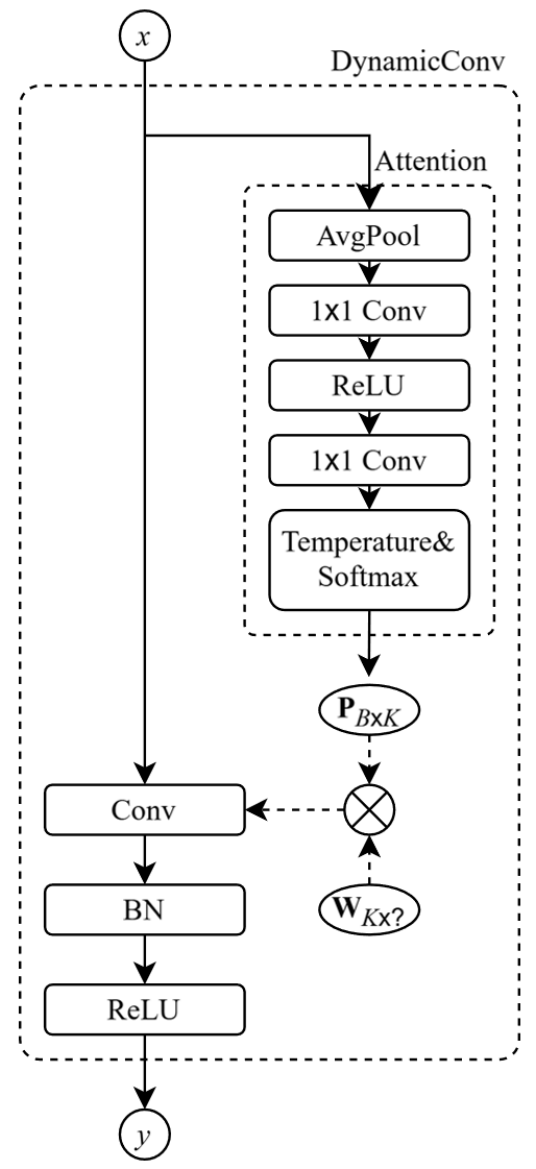}}}
	(b){{\includegraphics[scale=0.5,angle=0]{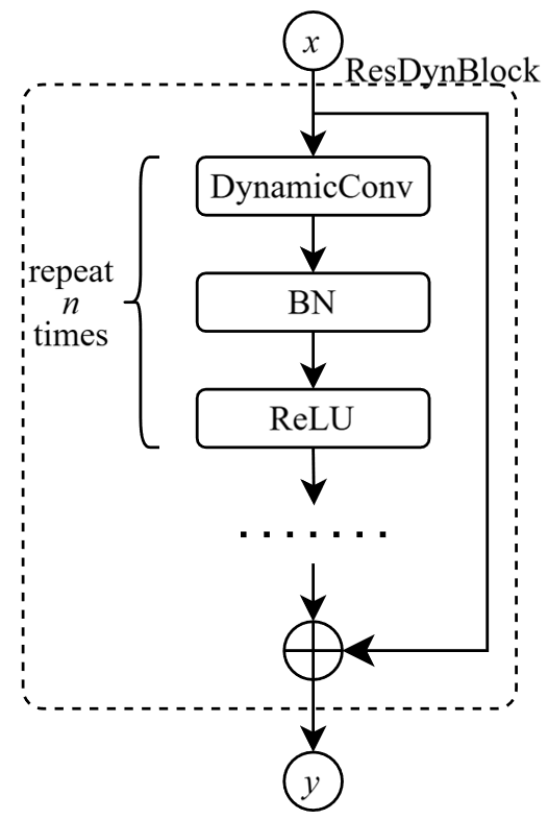}}}
	\caption{Fundamental building block of ResDynUNet++. (a) Dynamic convolution module: An attention module computes weights $\pi_k$ to aggregate K static kernels into a single dynamic kernel for each input sample. (b) Residual dynamic convolution block: A series of dynamic convolution, batch normalization, and ReLU layers are stacked, with a residual connection from the input to the output of the block.}
	\label{fig_resdynblock}
\end{figure}

The output of the dynamic perceptron is given by:
\begin{equation}
	y =\sigma(\tilde{\bm{W}}(\bm{x})\bm{x} + \tilde{\bm{b}}(\bm{x}))\,,
\end{equation}
where the aggregated weight $\tilde{\bm{W}}(\bm{x})$ and bias $\tilde{\bm{b}}(\bm{x})$ are defined as:
\begin{equation}
	\tilde{\bm{W}}(\bm{x}) = \sum_{k=1}^{K} \pi_k(\bm{x})\tilde{\bm{W}}_k, \quad \tilde{\bm{b}}(\bm{x}) = \sum_{k=1}^{K} \pi_k(\bm{x})\tilde{\bm{b}}_k\,,
\end{equation}
subject to the constraints on the attention weights:
\begin{equation*}
	0 \le \pi_k(\bm{x}) \le 1, \quad \sum_{k=1}^{K} \pi_k(\bm{x}) = 1 \,.
\end{equation*}
A distinguishing feature of the dynamic perceptron is that the attention weights $\{\pi_k(\bm{x})\}$ are input-adaptive rather than static. These weights determine the optimal aggregation of the linear experts $\{\tilde{\bm{W}}_k \bm{x} + \tilde{\bm{b}}_k\}$ for a specific input. Formally, $\{\pi_k(\bm{x})\}$ are computed using a softmax function with a temperature parameter $T$, which controls the sharpness of the distribution:
\begin{equation}
	\pi_k(\bm{x}) = \frac{\exp(\alpha_k(\bm{x})/T)}{\sum_{j=1}^K \exp(\alpha_j(\bm{x})/T)},
\end{equation}
where $\alpha_j(\bm{x})$ represents the attention logit for the $j$-th expert. \Fref{fig_resdynblock}\,(a) illustrates the structure of the dynamic convolution module.

\paragraph{Residual dynamic convolution block (ResDynBlock).}
The fundamental building block of our network is the ResDynBlock, illustrated in \Fref{fig_resdynblock}\,(b). This block comprises a Dynamic Convolution layer, followed by Batch Normalization (BN) and a Rectified Linear Unit (ReLU) activation. A residual connection is added from the input of the block to its output. This residual structure helps prevent vanishing gradients and allows for the training of deeper networks. In our implementation, the number of parallel kernels, K, in the Dynamic Convolution layer is set to 2.

\begin{figure}[htbp!]
	\centering
	\includegraphics[scale=0.5,angle=0]{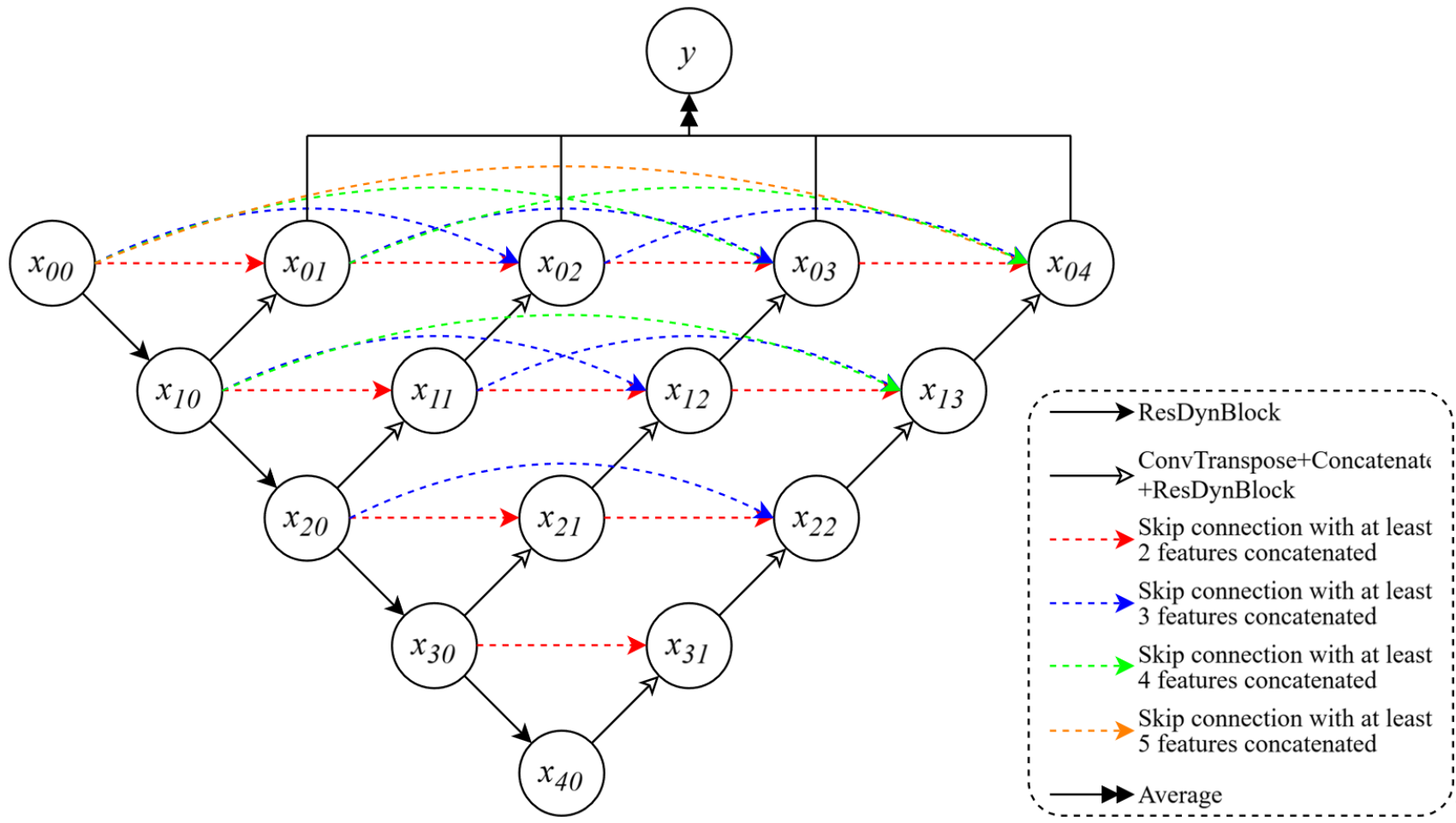}
	\caption{Overall architecture of ResDynUNet++.}
	\label{fig_resdynunetpp}
\end{figure}

\begin{figure}[htbp!]
		\centering
		{{\includegraphics[scale=0.5,angle=0]{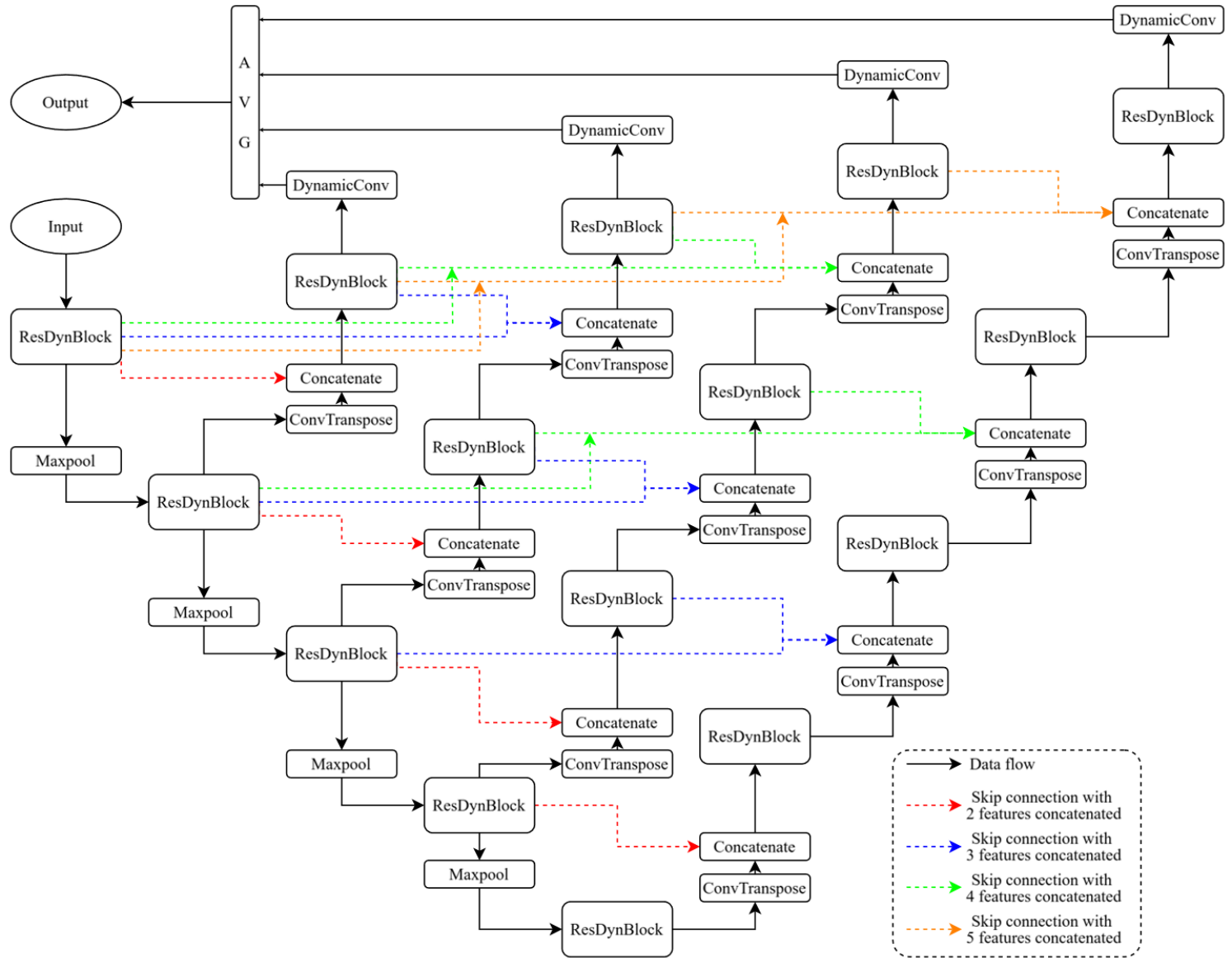}}}
		\caption{Detailed structure of ResDynUNet++.}
		\label{fig_detailed_struc}
	\end{figure}

\paragraph{Network architecture.}
\Fref{fig_resdynunetpp} shows the overall architecture of the proposed ResDynUNet++. Built upon the backbone of UNet++, ResDynUNet++ replaces each standard convolution layer with a residual dynamic block (ResDynBlock). The architecture features a deeply supervised encoder-decoder network with nested, dense skip pathways. The skip pathways connect feature maps from the encoder to the decoder at multiple semantic levels, which allows the model to learn from features of varying complexity. The final output is an aggregation of outputs from different levels of the decoder, which further improves performance. A more detailed view of the ResDynUNet++ structure, explicitly depicting the constituent blocks and operations, is provided in \Fref{fig_detailed_struc}.

\subsection{Training of the reconstruction operator} \label{SecTraining}
The complete reconstruction operator $\mathcal{A}_{\Theta}^{\dagger}$ (equation \eref{eq:Full_progress}) is trained to find its optimal parameters $\Theta$ by minimizing a supervised loss function. 
Let $\mathcal{D}_{\rm train} = \{(\mathbf{p}^{(s)}, \mathbf{y}^{(s)})\}_{s=1}^N$ denote the training set of input-output pairs , where $\mathbf{p}^{(s)}=(\mathbf{p}_1^{(s)}, \mathbf{p}_2^{(s)})$ is the measured projection data and $\mathbf{y}^{(s)} = (\mathbf{f}^{*}_{s}, \mathbf{g}^{*}_{s})$ is the corresponding two-channel ground-truth image. For each sample from the training set, the forward pass of the reconstruction operator begins by applying the fixed OPMT iterations $\mathcal{F}^n$ to the projection data $\mathbf{p}^{(s)}$, and the resulting intermediate solution is then passed to the learnable network $\Lambda_{\Theta}$ to produce the final prediction. The number of iterations $n$ is treated as a hyper-parameter, e.g., we set $n=10$.
The loss function $\mathcal{L}(\Theta)$ is defined as the Mean Squared Error (MSE) between the network's prediction $({\mathbf{f}}, {\mathbf{g}})$ and the ground-truth image $(\mathbf{f}^{*}, \mathbf{g}^{*})$:
\begin{equation}
	\mathcal{L}(\Theta) = \frac{1}{|\mathcal{D}_{\rm train}|} \sum_{s \in \mathcal{D}_{\rm train}} \frac{1}{2}\left( \mathrm{MSE}({\mathbf{f}}_s, \mathbf{f}_{s}^{*}) + \mathrm{MSE}({\mathbf{g}}_s, \mathbf{g}_{s}^{*}) \right),
\end{equation}
where $({\mathbf{f}}_s, {\mathbf{g}}_s) = \Lambda_{\Theta}(\mathcal{F}^n(\mathbf{p}^{(s)}))$. The minimization of the loss function is performed iteratively using the Adam optimizer. The training proceeds in epochs, where one epoch constitutes a full pass over the entire training set $\mathcal{D}_{\rm train}$. The data is processed in mini-batches of a predefined size. In the Dynamic Convolution layers, the temperature parameter $T$ is initialized at $T_0$ and annealed over the course of training to a minimum value, $T_{\min}$.

	\section{Experiments and results}
	
	\subsection{Experimental setup}	
\begin{figure}[htbp!]
		\centering
		(a){{\includegraphics[scale=0.45,angle=0]{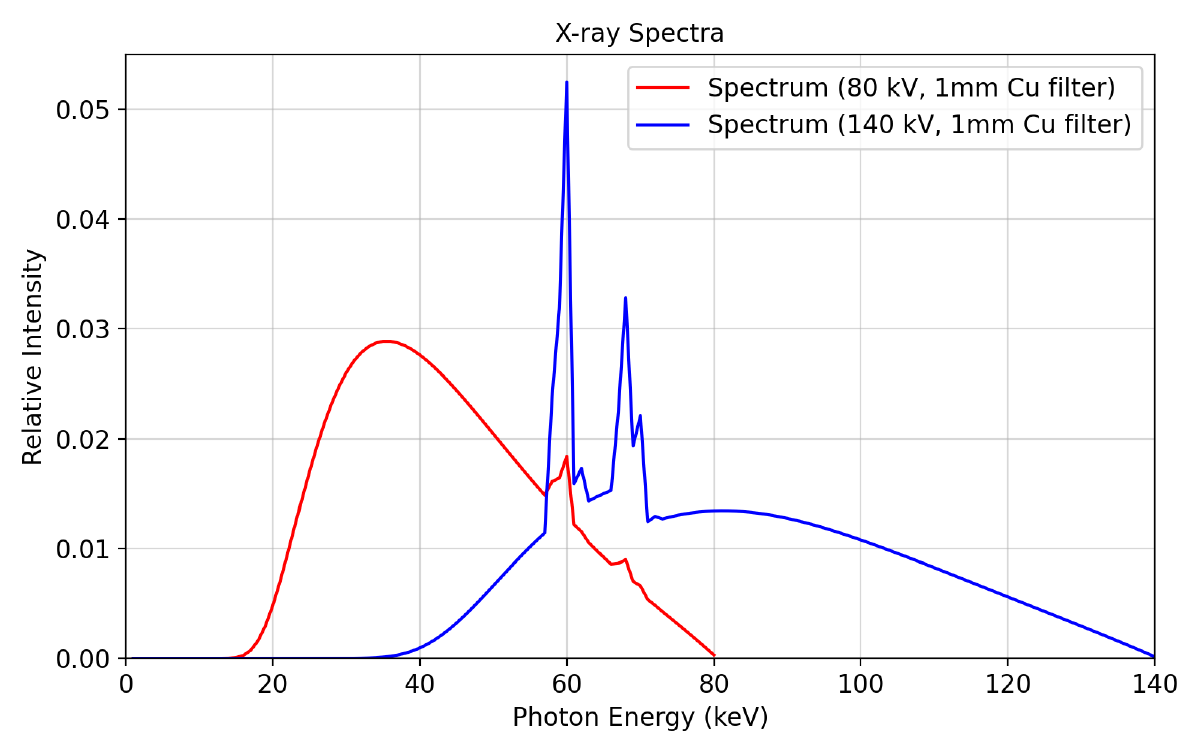}}}
		(b){{\includegraphics[scale=0.45,angle=0]{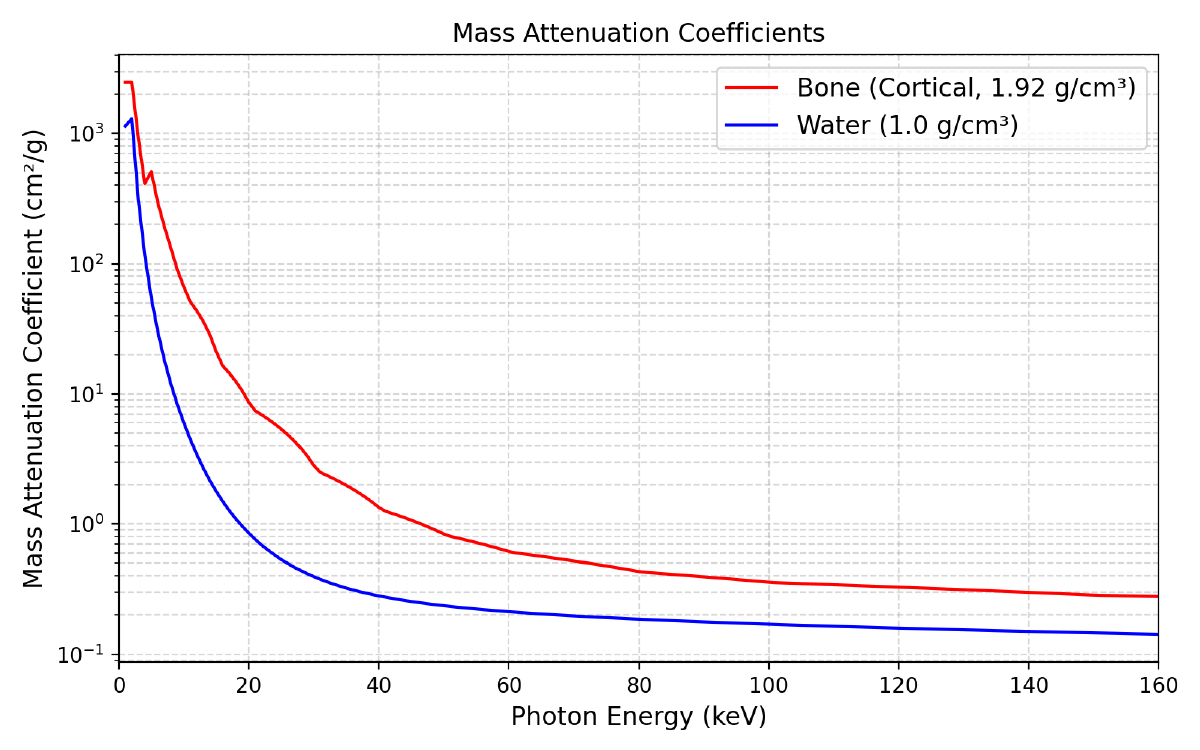}}}
		\caption{Experimental setup. (a) X-ray spectra. (b) Mass attenuation coefficients.}
		\label{fig_experimental setup}
	\end{figure}

	The X-ray spectra for the dual-energy simulation are generated using the SpectrumGUI software (http://spectrumgui.sourceforge.net/). Two distinct spectra are produced: $S_1(E)$ at tube voltage 80 kV, and $S_2(E)$ at 140 kV. Both incorporated a 1 mm copper filter and are calculated with a 1 keV energy resolution, as illustrated in \Fref{fig_experimental setup}\,(a). The mass attenuation coefficients for the basis materials, bone ($\phi(E)$) and water ($\theta(E)$), are also obtained from SpectrumGUI (\Fref{fig_experimental setup}\,(b)). The fan-beam CT geometry is defined by the following parameters: number of projection angles $n_S = 60$; number of detector elements $n_{D} = 256$; detector element size $l_D = 0.2$; source-to-object distance $D_1=490$; and object-to-detector distance $D_2=390$.
	
	The performance of all models will be evaluated quantitatively using three standard metrics: Mean Squared Error (MSE), Peak Signal-to-Noise Ratio (PSNR), and the Structural Similarity Index Measure (SSIM). Lower MSE and higher PSNR and SSIM values indicate better reconstruction quality.
	
	\subsection{Example 1: Phantom}
\begin{figure}[htbp!]
	 	\centering
	 	(a){\includegraphics[scale=0.4,angle=0]{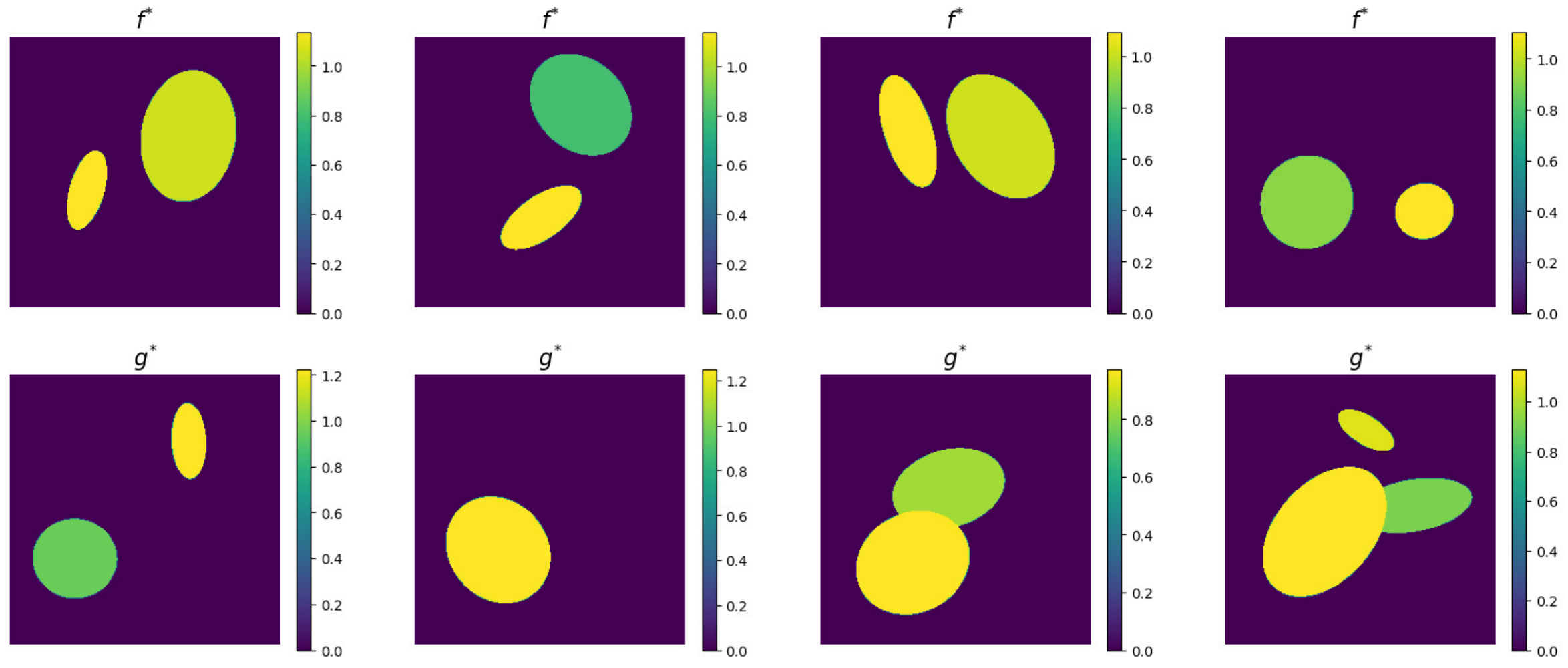}}
	 	(b){\includegraphics[scale=0.4,angle=0]{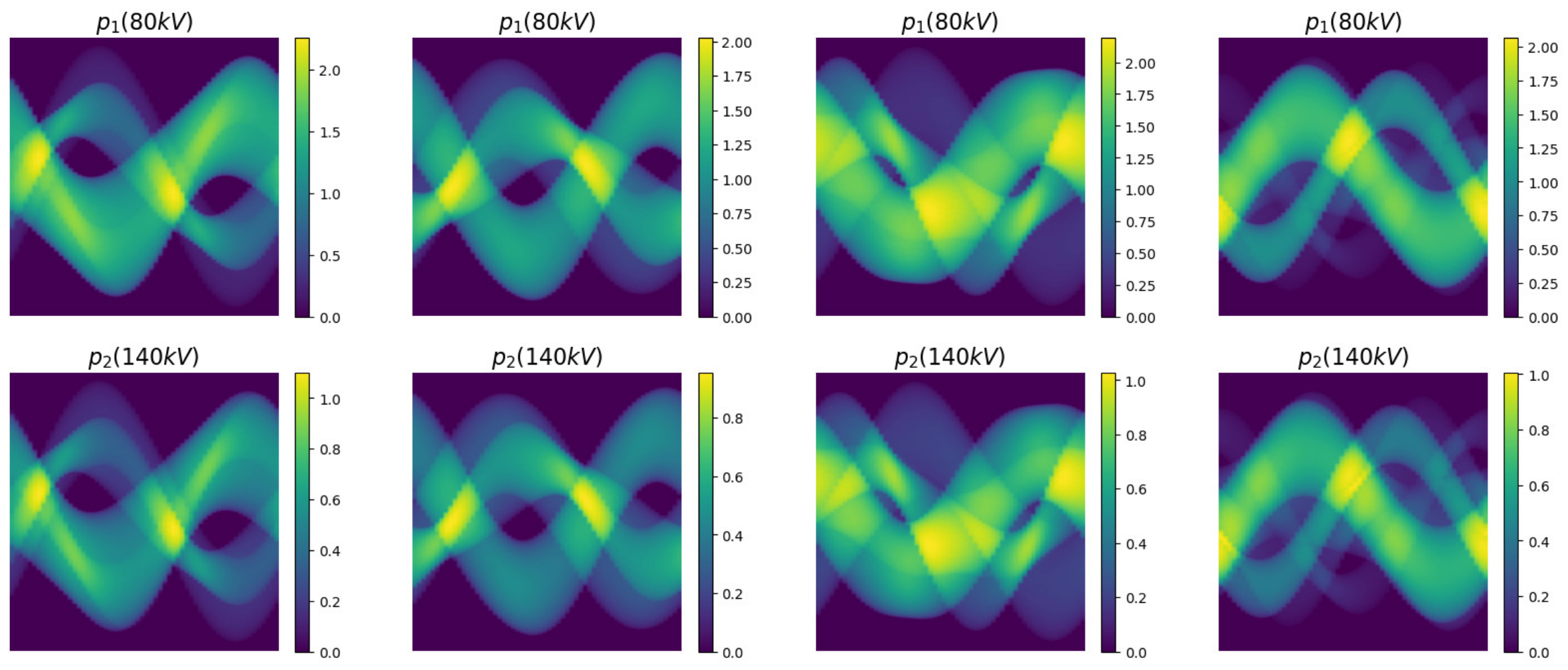}}
	 	\caption{Synthetic phantoms and projection data. (a) 4 of 3000 pairs in the training set. Each column represents one pair of phantoms, with the first and second rows showing the bone- and water-basis density images, $\mathbf{f}^{*}$ and $\mathbf{g}^{*}$, respectively. (b) Corresponding projection data. Poisson noise is introduced following equation (\ref{PoissonNoise}). The top row shows the low-energy spectra $\mathbf{p}_1$, and the bottom row shows the high-energy spectra $\mathbf{p}_2$.} 
	 	\label{fig_phantom_dataset}
	 \end{figure}

	We first evaluate the proposed method on a simulated phantom dataset. A total of 3500 pairs of phantom images (256$\times$256 pixels) are generated, split into training, validation, and test sets in a 3000:400:100 ratio. Each phantom contains a random number of ellipses, drawn from a Poisson distribution ($\lambda=2$). The intensity of each ellipse is sampled from a Gaussian distribution ($\mu=1, \sigma=0.1$), with overlapping regions assigned the maximum intensity of the constituent ellipses. Representative examples from the training set are shown in \Fref{fig_phantom_dataset}\,(a).
	 
	 The projection data are generated according to equation (\ref{eq:projection}) by adding Poisson noise,
\begin{equation} \label{PoissonNoise}
p_{k,\rm{noisy}}=-\ln \left(\frac{{\rm{Poissrnd}}(I_0\,e^{-p_k})}{I_0}\right)\,,\quad k=1,2
\end{equation}
where $p_{k,\rm{noisy}}$ and $p_k$ denote the projection data with and without noise, respectively. In equation (\ref{PoissonNoise}), $I_0$ indicates the X-ray intensity for each path, and ${\rm{Poissrnd}}(I_0\,e^{-p_k})$ generates random numbers from the Poisson distribution with mean $I_0\,e^{-p_k}$; we set $I_0=10^5$ to simulate the situation of low-dose CT. \Fref{fig_phantom_dataset}\,(b) plots the projection data for the 4 pairs of synthetic phantoms displayed in \Fref{fig_phantom_dataset}\,(a).
	 
\begin{figure}[htbp!]
		\centering
		\includegraphics[scale=0.4,angle=0]{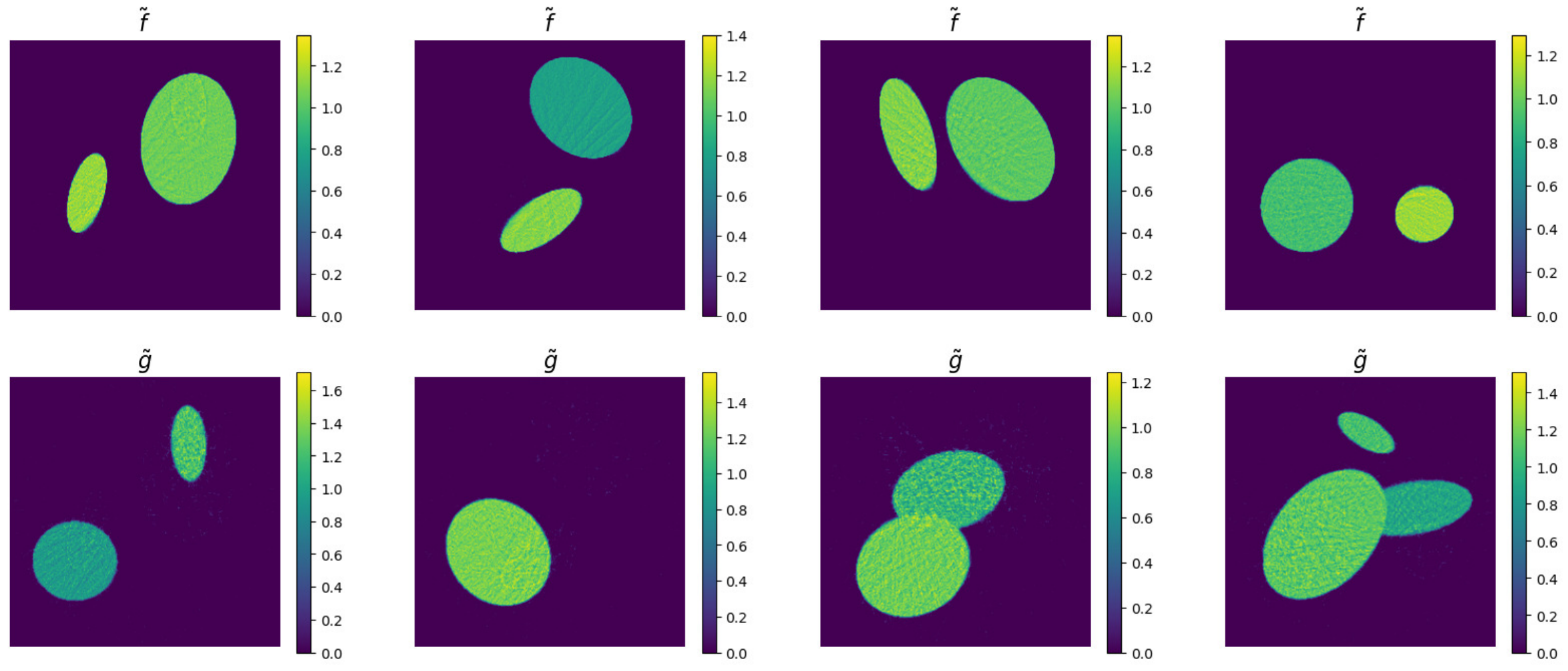}
		\caption{OPMT intermediate solutions for the 4 pairs of synthetic phantoms displayed in \Fref{fig_phantom_dataset}. $(\mathbf{\tilde{f}}, \mathbf{\tilde{g}}) = \mathcal{F}^n(\mathbf{p})$, where the number of iterations $n$ is a fixed hyper-parameter, and we set it as $n=10$. The top row shows the bone-basis density $\tilde{\mathbf{f}}$, and the bottom row shows the water-basis density $\tilde{\mathbf{g}}$.}
		\label{fig_intersolve_phantom}
	\end{figure}

\begin{figure}[htbp!]
		\centering
		\includegraphics[scale=0.65,angle=0]{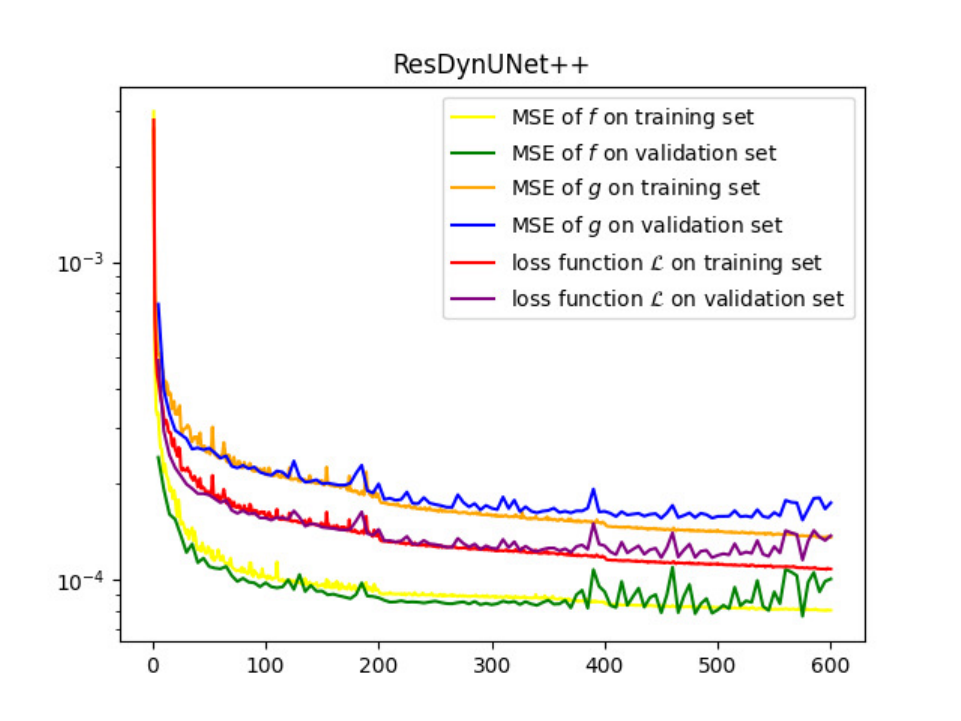}
		\caption{Convergence plot during training on the phantom dataset.}
		\label{fig_tr_phantom}
	\end{figure}

The OPMT iterations then produce intermediate material-decomposed images from the noisy projection data. As shown in \Fref{fig_intersolve_phantom}, these intermediate reconstructions, while correctly separating the basis materials, suffer from significant noise and artifacts, highlighting the need for a refinement step. 
The intermediate reconstruction is passed through the ResDynUNet++ model $\Lambda_{\Theta}$ to yield the final solution. The network parameters $\Theta$ are optimized using the training strategy detailed in Section \ref{SecTraining}. \Fref{fig_tr_phantom} shows the convergence plot in the training process, where we further include the MSE of the two-channel outputs, $\mathbf{f}$ and $\mathbf{g}$, respectively. It shows that the three curves (MSE of $\mathbf{f}$, MSE of $\mathbf{g}$, and total loss $\mathcal{L}$) converge in the same manner, and their overfitting appears around the same number of iterations. It implies that channel imbalance is insignificant for our ResDynUNet++ model. 

\begin{figure}[htbp!]
		\centering
		(a){{\includegraphics[scale=0.5,angle=0]{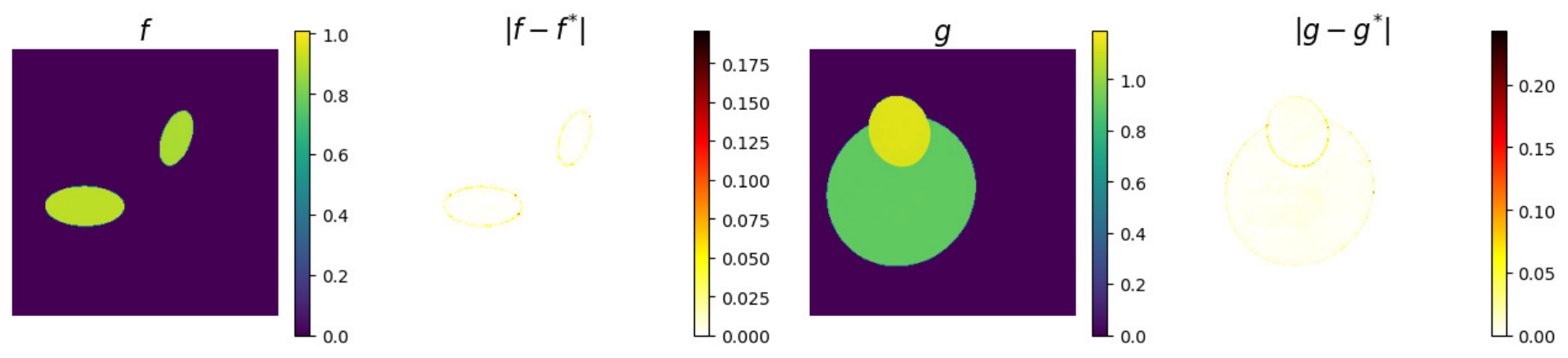}}}
		(b){{\includegraphics[scale=0.5,angle=0]{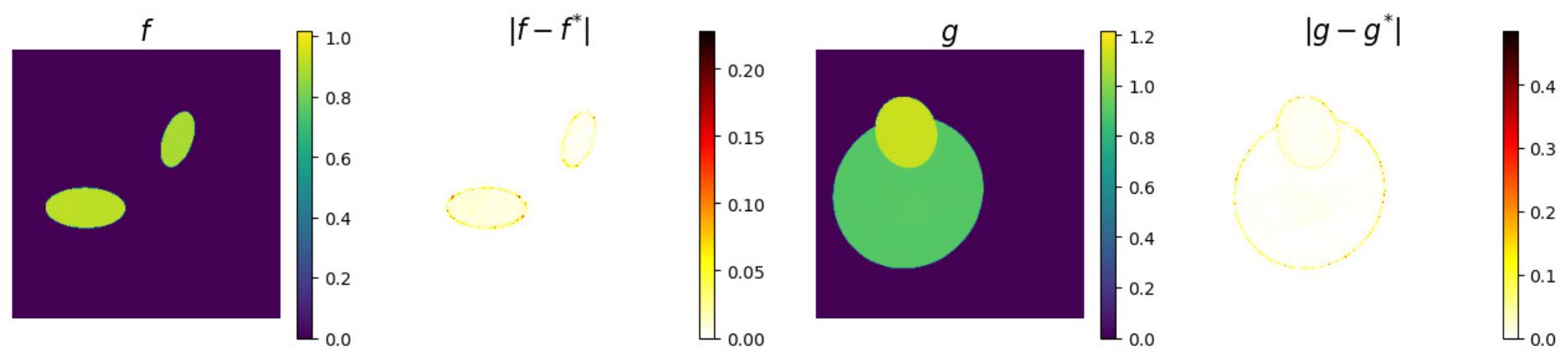}}}
		(c){{\includegraphics[scale=0.5,angle=0]{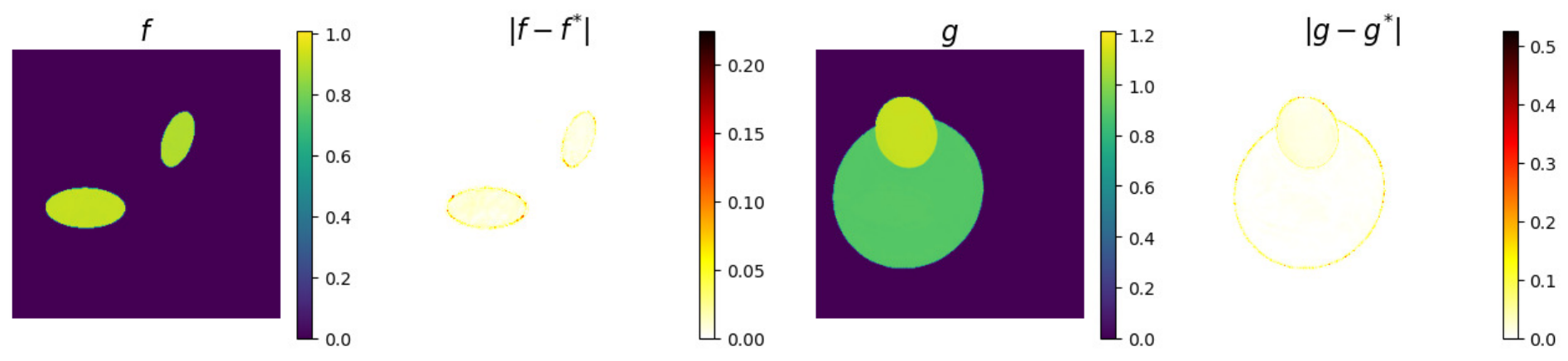}}}
		\caption{Prediction results for a sample from the test set. The predicted bone-basis ($\mathbf{f}$) and water-basis ($\mathbf{g}$) density images are shown, along with their absolute difference maps against the ground truth ($\mathbf{f}^{*}$, $\mathbf{g}^{*}$). Results from: (a) ResDynUNet++; (b) DynUNet++; (c) UNet++.}
		\label{fig_res_phantom}
	\end{figure}

After training, the complete reconstruction operator $\mathcal{A}_{\Theta}^{\dagger}$ is applied to the test set. \Fref{fig_res_phantom} illustrates the qualitative reconstruction, presenting the result for one sample from the test set. To demonstrate the performance improvement, we compare the prediction results using ResDynUNet++, DynUNet++, and UNet++ for the data-driven part. Here, DynUNet++ denotes the architecture without residual connection in the dynamic convolution block. Visually, ResDynUNet++ produces images that are remarkably cleaner and structurally more accurate than the OPMT intermediate solutions, and the results outperform those from DynUNet++ and UNet++. In \Tref{tab:phantom_results}, we report the average MSE, PSNR and SSIM values on 100 pairs of test samples. The quantitative results confirm the superiority of ResDynUNet++, which achieves the best scores across all metrics (MSE, PSNR, and SSIM) for both basis materials, significantly outperforming the other models.
	
	\Table{\label{tab:phantom_results}Quantitative results (MSE, PSNR, SSIM) averaged over the 100-sample test set}
	\br
	Metric & ResDynUNet++ & DynUNet++ & UNet++ \\
	\mr
	Average MSE (Bone) & \textbf{2.770e-5} & 6.860e-5 & 7.937e-5 \\
	Average PSNR (Bone) (dB) & \textbf{48.43} & 44.19 & 43.29 \\
	Average SSIM (Bone) & \textbf{0.999900} & 0.999742 & 0.999673 \\
	\mr
	Average MSE (Water) & \textbf{4.692e-5} & 2.377e-4 & 2.791e-4 \\
	Average PSNR (Water) (dB) & \textbf{45.97} & 37.89 & 37.04 \\
	Average SSIM (Water) & \textbf{0.999806} & 0.999068 & 0.998866 \\
	\br
	\endTable
	
	\subsection{Example 2: Clinical head CT}
	
\begin{figure}[htbp!]
		\centering
		(a){\includegraphics[scale=0.4,angle=0]{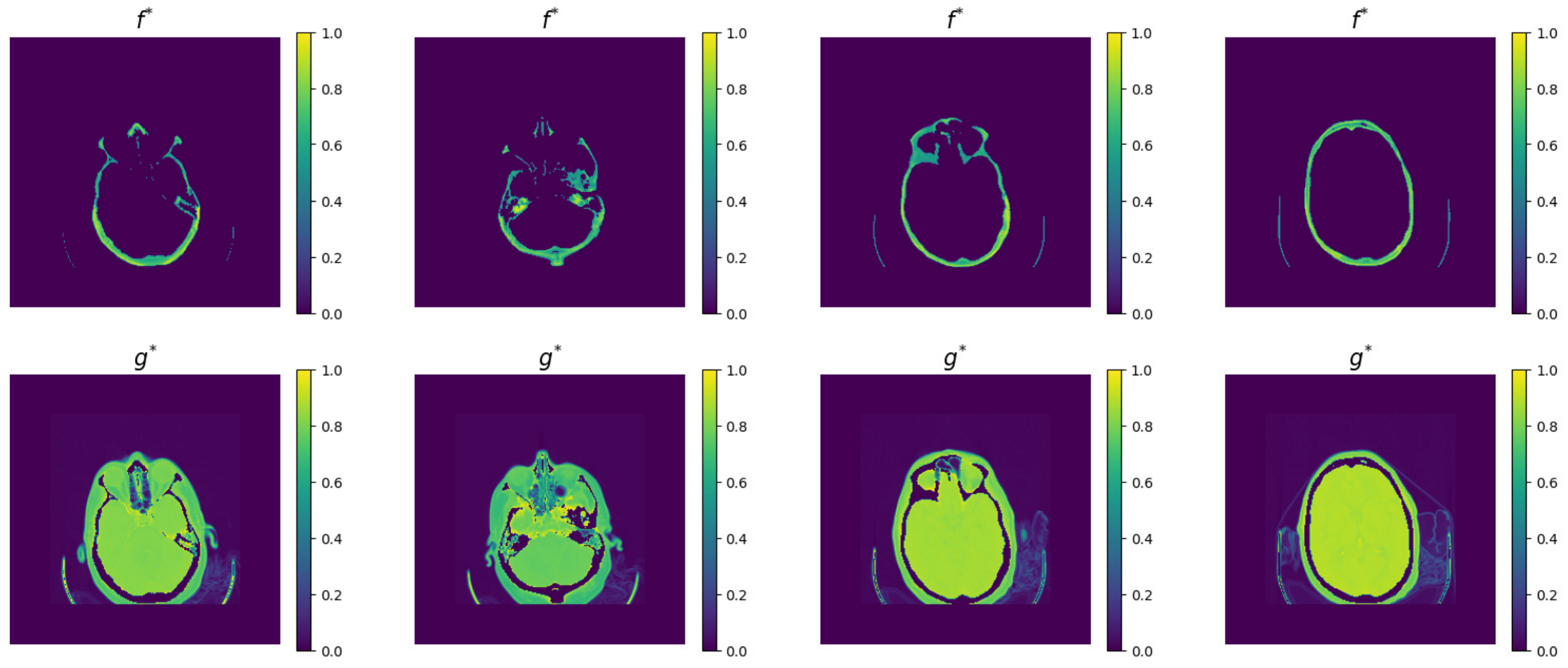}}
		(b){\includegraphics[scale=0.4,angle=0]{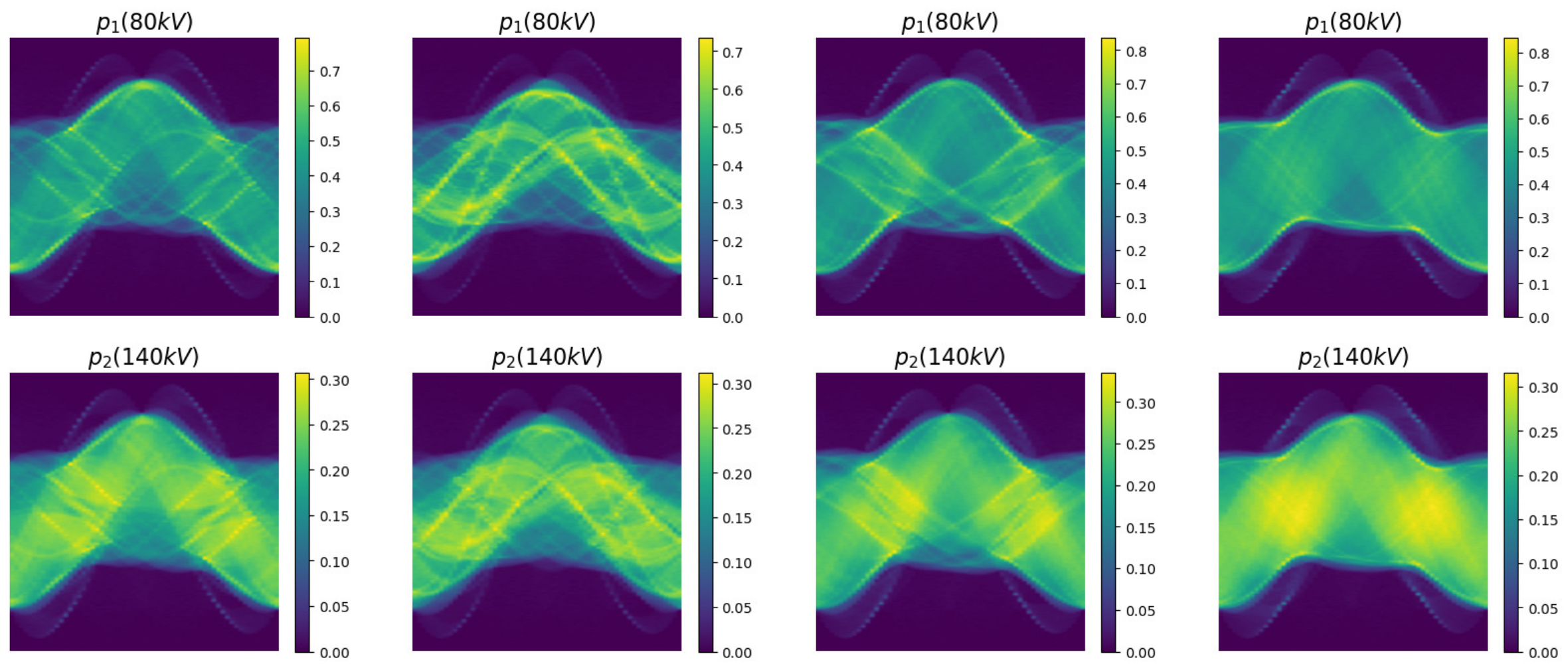}}
		\caption{Clinical dataset adapted from the public head CT collection CQ500. (a) 4 of 800 pairs of head scans in the training set. Each column represents one pair of head images, with the first and second rows showing the bone- and water-basis density maps, $\mathbf{f}^{*}$ and $\mathbf{g}^{*}$, respectively. (b) Corresponding projection data, with the top and bottom rows showing the low-energy ($\mathbf{p}_1$) and high-energy ($\mathbf{p}_2$) spectra, respectively.}
		\label{fig_real_dataset}
	\end{figure}

	This study utilizes a clinical head CT dataset consisting of 1000 scans (256$\times$256 pixels per slice). The dataset is adapted from the public head CT collection CQ500  (https://public.md.ai/hub/projects/public), which is licensed under CC BY-NC-SA 4.0. The data is partitioned into training, validation, and test sets in an 8:1:1 ratio. Representative ground truth samples from the training set are displayed in \Fref{fig_real_dataset}\,(a), and the corresponding projection data are illustrated in \Fref{fig_real_dataset}\,(b).
	
\begin{figure}[htbp!]
		\centering
		\includegraphics[scale=0.4,angle=0]{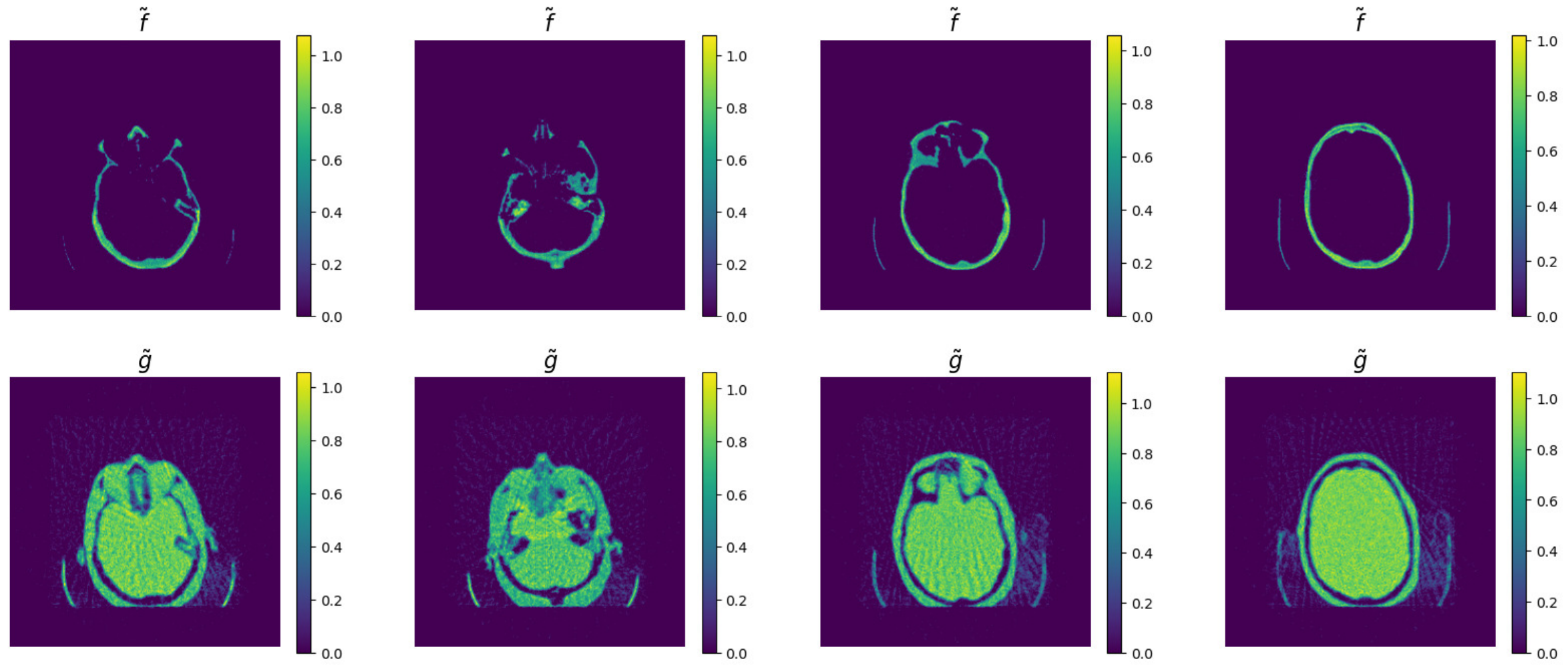}
		\caption{OPMT intermediate solutions for the 4 pairs of head images displayed in \Fref{fig_real_dataset}. $(\mathbf{\tilde{f}}, \mathbf{\tilde{g}}) = \mathcal{F}^n(\mathbf{p})$, where the number of iterations $n$ is a fixed hyper-parameter, and we set it as $n=10$. The top row shows the bone-basis density $\tilde{\mathbf{f}}$, and the bottom row shows the water-basis density $\tilde{\mathbf{g}}$.}
		\label{fig_intersolve_real}
	\end{figure}

\begin{figure}[htbp!]
		\centering
		{{\includegraphics[scale=0.5,angle=0]{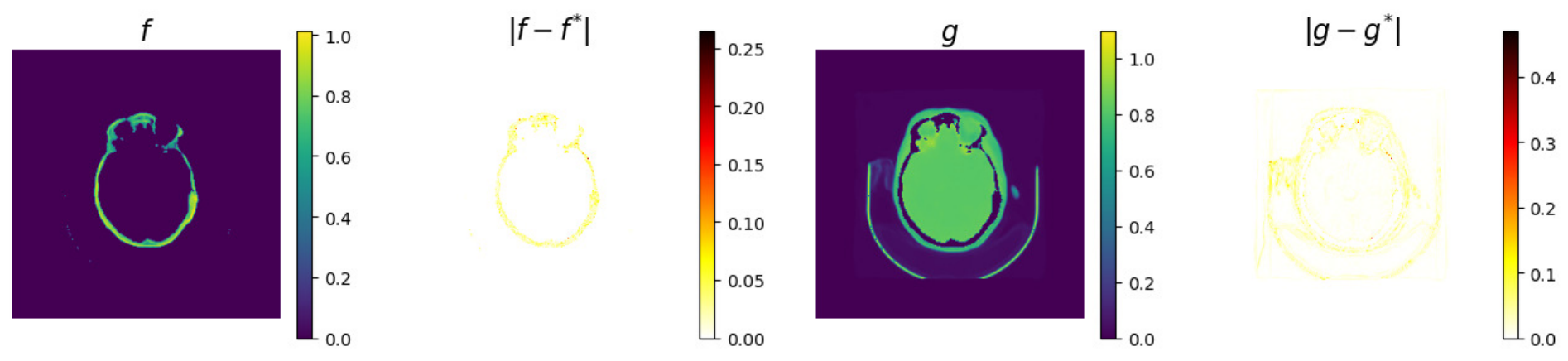}}}
		{{\includegraphics[scale=0.5,angle=0]{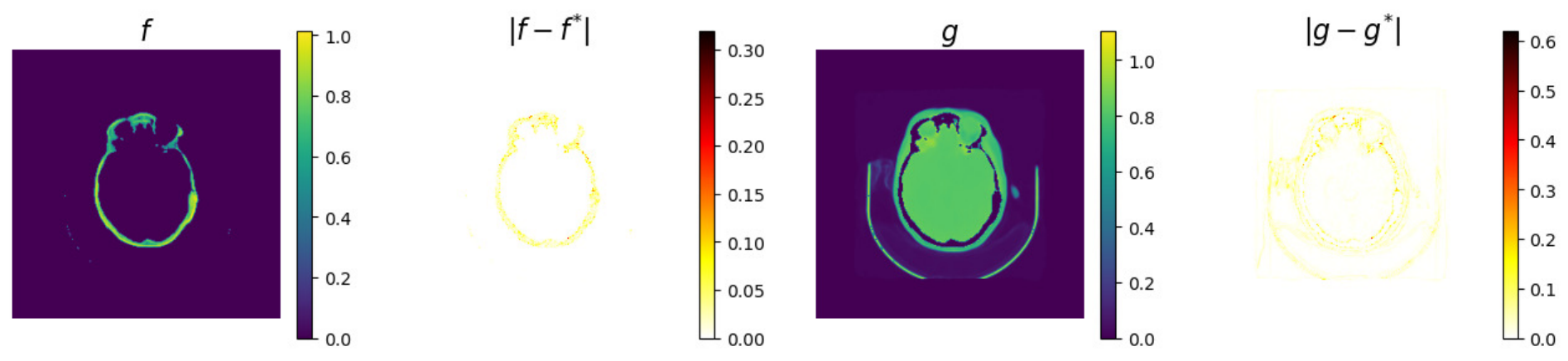}}}
		{{\includegraphics[scale=0.5,angle=0]{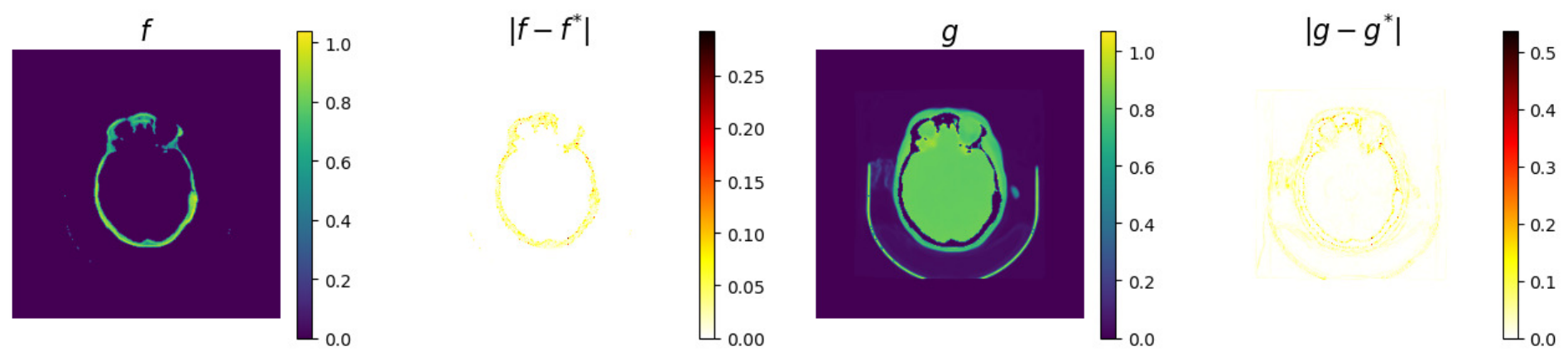}}}
		\caption{Prediction results for a sample from the test set. The predicted bone-basis ($\mathbf{f}$) and water-basis ($\mathbf{g}$) density maps are shown, along with their absolute difference maps against the ground truth ($\mathbf{f}^{*}$, $\mathbf{g}^{*}$). Results from: (a) ResDynUNet++; (b) DynUNet++; (c) UNet++.}
		\label{fig_res_real}
	\end{figure}

	The reconstruction operator $\mathcal{A}_{\Theta}^{\dagger}$ is initialized using OPMT iterations $\mathcal{F}^n$, where the iteration number $n$ is a hyper-parameter set to $n=10$. \Fref{fig_intersolve_real} shows some examples of the OPMT intermediate solutions. These solutions exhibit a reasonable decomposition of basis materials, demonstrating the effect of the model-driven part, but suffer from contamination of noise and artifacts. The data-driven part $\Lambda_{\Theta}$ is then trained to improve the reconstruction.	
	The trained framework is evaluated on the 100-sample test set. \Fref{fig_res_real} shows the prediction result for one sample from the test set, and \Tref{tab:real_results} reports the average values of MSE, PSNR and SSIM on 100 pairs of test samples. To demonstrate the performance improvement, we compare the prediction results using ResDynUNet++, DynUNet++, and UNet++ for the data-driven part.
	The results validate the effectiveness of our approach. The visual quality of the bone and water density maps is substantially improved after refinement with ResDynUNet++. Quantitatively, our proposed model consistently achieves the lowest MSE and the highest PSNR and SSIM values, demonstrating its robust performance on complex, real clinical data.
	
	\Table{\label{tab:real_results}Quantitative results (MSE, PSNR, SSIM) averaged over the 100-sample test set}
	\br
	Metric & ResDynUNet++ & DynUNet++ & UNet++ \\
	\mr
	Average MSE (Bone) & \textbf{5.487e-5} & 8.495e-5 & 9.792e-5 \\
	Average PSNR (Bone) (dB) & \textbf{43.90} & 41.77 & 41.05 \\
	Average SSIM (Bone) & \textbf{0.998094} & 0.996938 &  0.996525 \\
	\mr
	Average MSE (Water) & \textbf{3.471e-4} & 5.062e-4 & 5.463e-4 \\
	Average PSNR (Water) (dB) & \textbf{35.40} & 33.51 & 33.09 \\
	Average SSIM (Water) & \textbf{0.997819} & 0.996835 & 0.996575 \\
	\br
	\endTable
	
	\section{Conclusions}
	We propose a hybrid two-stage reconstruction operator, $\mathcal{A}_{\Theta}^{\dagger}$, that effectively combines a classical iterative algorithm with a novel deep learning model for dual-spectral CT. The oblique projection modification technique (OPMT) is selected as the model-driven component of $\mathcal{A}_{\Theta}^{\dagger}$. Due to its fast convergence, the OPMT rapidly generates an intermediate solution that achieves successful basis material decomposition, a challenging task for purely data-driven approaches. 
	To refine this intermediate solution, which is typically corrupted by noise and artifacts, we develop ResDynUNet++ as the data-driven component of $\mathcal{A}_{\Theta}^{\dagger}$. This novel deep neural network integrates the multi-scale feature fusion of UNet++, the sample-adaptive capabilities of dynamic convolution, and the stable training provided by residual connections. This architecture is specifically designed to overcome challenges such as channel imbalance and large artifacts near interface regions in dual-spectral CT reconstruction, yielding clean and accurate final solutions. Extensive experiments conducted on both synthetic and clinical CT data validate the superiority of our model over UNet++ and its variants. The results highlight the potential of our proposed framework for solving challenging medical imaging problems in dual-spectral CT.

	\section*{Acknowledgments}
	Wenbin Li is supported by the Natural Science Foundation of Shenzhen (JCYJ20240813104841055), and the Fundamental Research Funds for the Central Universities (HIT.OCEF.2024017). Shusen Zhao is supported by Shenzhen Science and Technology Program (Grant No. JSGGZD20220822095600001), and the Longhua District Science and Innovation Commission Project Grants of Shenzhen (Grant No.  20250113G43468522).
	
\newpage
\section*{References}
\bibliographystyle{plain}

\end{document}